\documentclass[letterpaper, 10 pt, conference]{ieeeconf}

\IEEEoverridecommandlockouts
\makeatletter
\let\NAT@parse\undefined
\makeatother
\usepackage[colorlinks, citecolor=green]{hyperref}
\usepackage{booktabs}
\usepackage{multirow}
\usepackage{siunitx}
\usepackage{graphicx}
\usepackage{fontawesome}
\usepackage{utfsym}
\usepackage{stfloats}
\usepackage{makecell}
\usepackage{amsmath}
\usepackage{amssymb}

\title
{
    \LARGE
    \bf
    Adapting Depth Anything to Adverse Imaging Conditions with Events
}

\author
{
    Shihan Peng, Yuyang Xiong, Hanyu Zhou, Zhiwei Shi, Haoyue Liu*, Gang Chen, Luxin Yan, and Yi Chang
    \thanks{*Corresponding author.}
    \thanks
    {
        Shihan Peng, Yuyang Xiong, Zhiwei Shi, Haoyue Liu, Luxin Yan, and Yi Chang are with the National Key Lab of Multispectral Information Intelligent Processing Technology, School of Artificial Intelligence and Automation, Huazhong University of Science and Technology, Wuhan, China. Email: {\tt\small \{pengshihan, xiongyuyang, shizhiwei, liuhy, yichang, yanluxin\}@hust.edu.cn}
    }
    \thanks
    {
        Hanyu Zhou is with the School of Computing, National University of Singapore. Email: {\tt\small hy.zhou@nus.edu.sg}
    }
    \thanks
    {
        Gang Chen is with the School of Computer Science and Engineering, Sun Yat-Sen University, Guangzhou 510275, China. Email: {\tt\small cheng83@mail.sysu.edu.cn}
    }
}

\begin{document}
    \maketitle
    \thispagestyle{empty} 
    \pagestyle{empty}

    
    \begin{abstract}
        Robust depth estimation under dynamic and adverse lighting conditions is essential for robotic systems. Currently, depth foundation models, such as Depth Anything, achieve great success in ideal scenes but remain challenging under adverse imaging conditions such as extreme illumination and motion blur. These degradations corrupt the visual signals of frame cameras, weakening the discriminative features of frame-based depths across the spatial and temporal dimensions. Typically, existing approaches incorporate event cameras to leverage their high dynamic range and temporal resolution, aiming to compensate for corrupted frame features. However, such specialized fusion models are predominantly trained from scratch on domain-specific datasets, thereby failing to inherit the open-world knowledge and robust generalization inherent to foundation models. In this work, we propose ADAE, an event-guided spatiotemporal fusion framework for Depth Anything in degraded scenes. Our design is guided by two key insights: 1) \textit{Entropy-Aware Spatial Fusion}. We adaptively merge frame-based and event-based features using an information entropy strategy to indicate illumination-induced degradation. 2) \textit{Motion-Guided Temporal Correction}. We resort to the event-based motion cue to recalibrate ambiguous features in blurred regions. Under our unified framework, the two components are complementary to each other and jointly enhance Depth Anything under adverse imaging conditions. Extensive experiments have been performed to verify the superiority of the proposed method. Our code will be released upon acceptance.
    \end{abstract}
    
    
    \section{INTRODUCTION}
\begin{figure}
        \centering
        \includegraphics[width=1.0\linewidth]{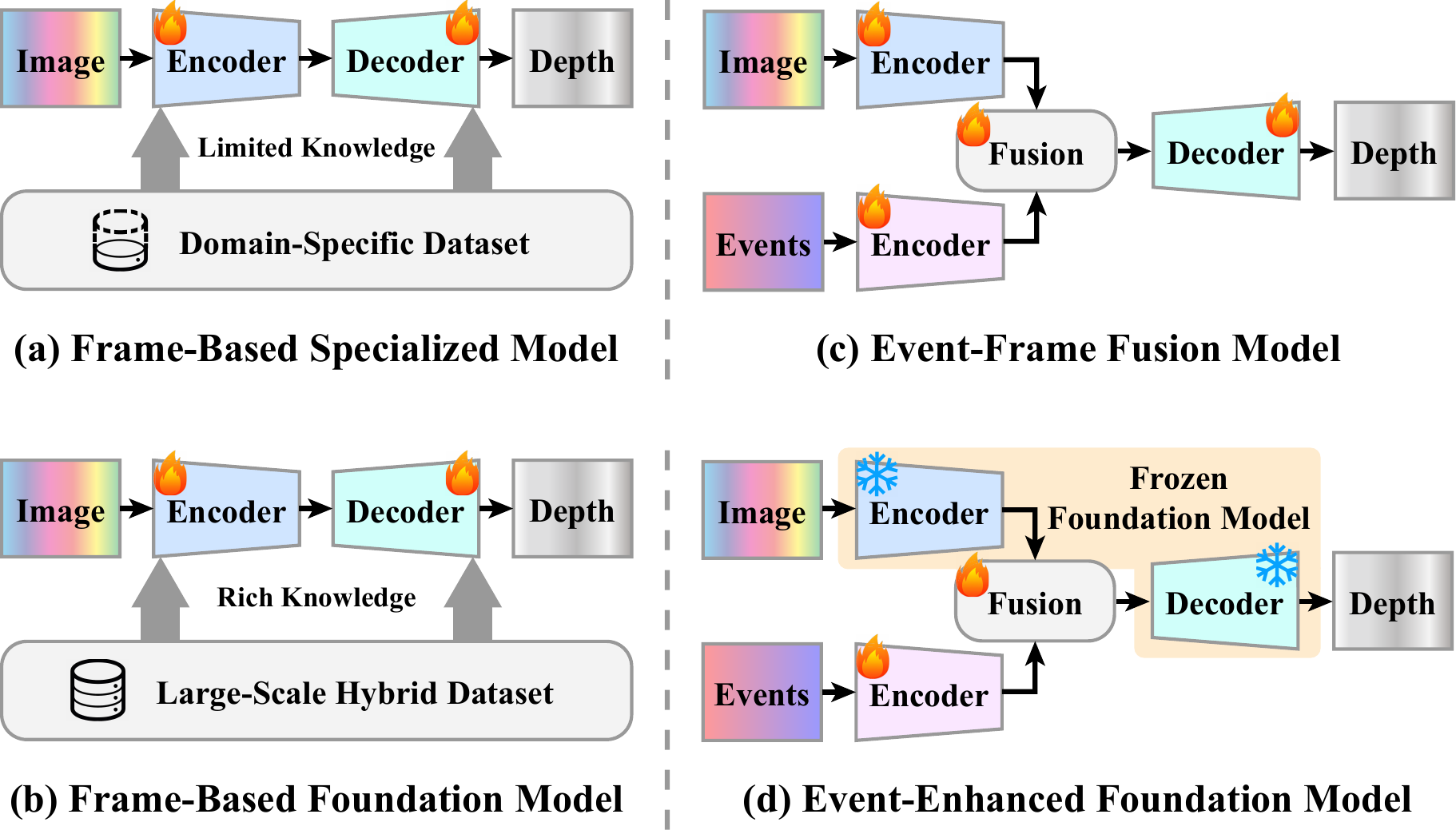}
        \caption
        {
            Comparison of four depth estimation paradigms. (a) Frame-based specialized models are trained on domain-specific datasets and suffer from poor generalization. (b) Frame-based foundation models leverage large-scale datasets for rich knowledge and strong generalization, but fail in adverse imaging conditions. (c) Event-frame fusion models enhance robustness but are also specialized and lack generalization. (d) Our event-enhanced foundation model synergizes the strong generalization of a frozen foundation model (b) with the signal-level robustness of event-based fusion (c).
        }
        \label{Fig:Paradigm_Comparison}
    \end{figure}

        Depth estimation serves as a cornerstone task for various vision applications, including augmented reality \cite{valentin2018depth}, autonomous driving \cite{zhou2023unsupervised}, and robotic perception \cite{dong2022towards}. Previous approaches \cite{eigen2014depth, wang2021regularizing, patil2022p3depth, zheng2023steps} rely on \textit{frame-based specialized models} (Figure \ref{Fig:Paradigm_Comparison}(a)) trained on domain-specific datasets, which struggle to generalize to unseen scenarios. Recently, \textit{frame-based foundation models} (Figure \ref{Fig:Paradigm_Comparison}(b)) \cite{ranftl2020towards, yin2023metric3d, hu2024metric3d, ke2024repurposing, gui2025depthfm}, represented by \textit{Depth Anything} \cite{yang2024depth}, have demonstrated remarkable zero-shot generalization across diverse scenarios by leveraging large-scale hybrid datasets. 
        
        While these frame-based approaches have achieved impressive performance in ideal environments, they remain vulnerable to adverse imaging conditions, particularly extreme illumination and motion blur. Extreme over- or underexposure leads to a loss of spatial information in image frames, while severe motion blur introduces temporal boundary ambiguity by blending foreground and background structures. In these situations, the fundamental visual signal captured by conventional cameras is intrinsically corrupted. This reveals a critical weakness of even the most powerful foundation models: their performance is ultimately constrained by the quality of the input signal. When information is irretrievably lost at the signal level, no amount of pre-trained knowledge can fully recover the non-existent structural details. 

        To address this signal-level challenge, a research avenue has focused on incorporating novel sensing modalities, leading to \textit{event-frame fusion models} (Figure \ref{Fig:Paradigm_Comparison}(c)) \cite{gehrig2021combining, shi2023even, xiao2024multimodal, duan2024fusing, pan2024srfnet, devulapally2024multi}. Event cameras, with their high dynamic range and high temporal resolution, can capture visual information in scenarios where frame-based cameras fail. By fusing event data with frames, these models can reconstruct depth in challenging scenes. However, similar to \textit{frame-based specialized models}, these fusion models are typically trained on domain-specific datasets, thus lacking the broad knowledge and superior generalization capability inherent to depth foundation models.

\begin{figure*}[t]
        \centering
        \includegraphics[width=1.0\textwidth]{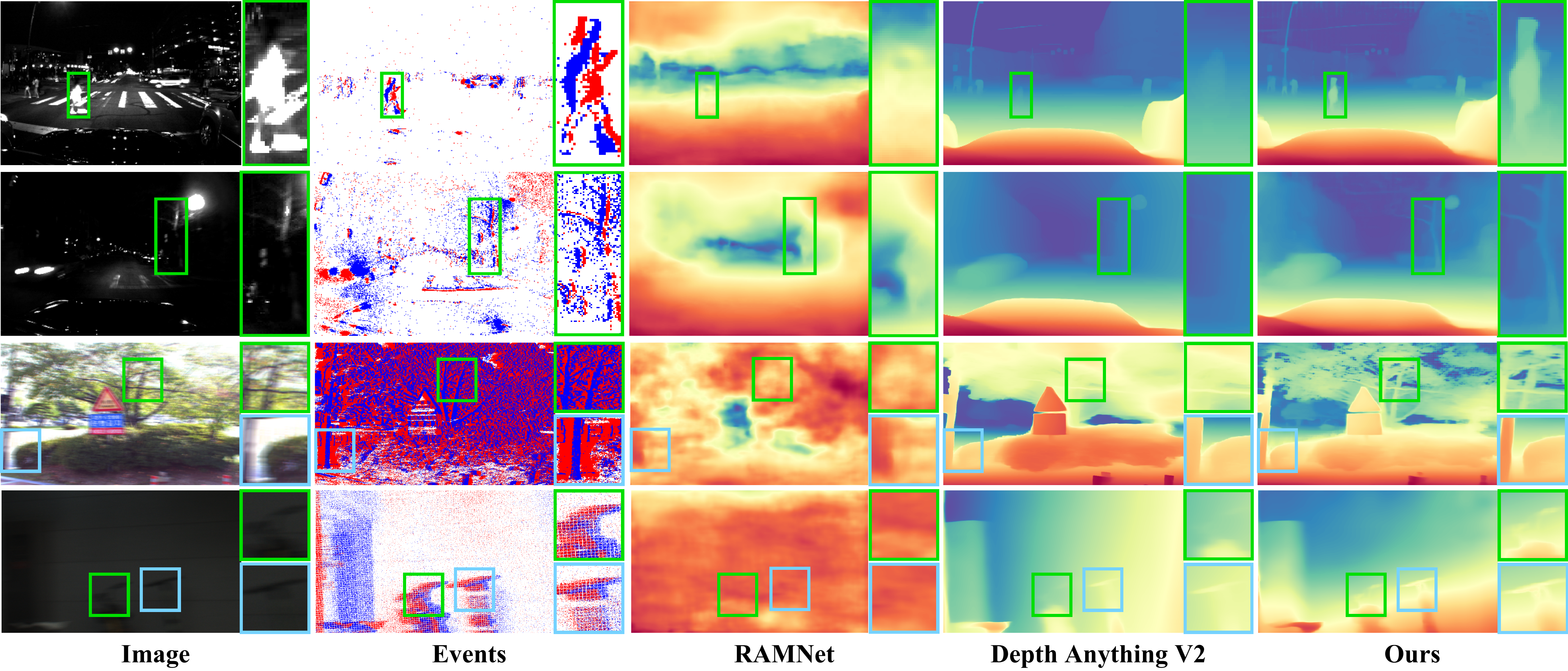}
        \caption
        {
            Zero-shot depth prediction on MVSEC \cite{zhu2018multivehicle}, EVRB \cite{kim2024cmta}, and NCER \cite{cho2023non} datasets. We compare our method (ADAE) with a representative event-frame fusion method (RAMNet) and the foundation model (Depth Anything V2). The top two rows show scenes with extreme illumination (over- and underexposure), while the bottom two rows contain motion blur. Our method produces more structurally complete and detailed depth maps.
        }
        \label{Fig:Model_Comparison}
    \end{figure*}

        In this work, we argue that data-driven generalization and event-based signal enhancement are complementary. To this end, we introduce the \textit{event-enhanced foundation model} (Figure \ref{Fig:Paradigm_Comparison}(d)) and propose ADAE, an event-guided spatiotemporal fusion framework. Instead of building a new network from scratch, ADAE synergizes a frozen Depth Anything model with event data via a cross-modal adapter. This allows our method to leverage event data to spatially compensate for information loss and temporally correct motion-induced blur, all while inheriting the generalization power of the depth foundation model. As illustrated in Figure \ref{Fig:Model_Comparison}, we enhance the robustness of the depth foundation model in degraded environments while preserving its generalization capability. Overall, our main contributions are summarized as follows:
        \begin{itemize}
            \item We present ADAE, an event-guided spatiotemporal fusion framework that integrates the resilience of event signals with a frozen foundation model to enhance robustness under extreme illumination and motion blur.

            \item We introduce \textit{Entropy-Aware Spatial Fusion} (\textbf{EASF}), which adaptively merges frame and event features using information entropy to correct illumination degradation.

            \item We introduce \textit{Motion-Guided Temporal Correction} (\textbf{MGTC}), which uses event-based optical flow to recalibrate blurred features and restore structural boundaries.
            
            \item We conduct extensive experiments on various datasets, and results demonstrate that ADAE enhances Depth Anything's performance under adverse imaging conditions while preserving its generalization capability.
        \end{itemize}

    \begin{figure*}[t]
        \centering
        \includegraphics[width=1.0\textwidth]{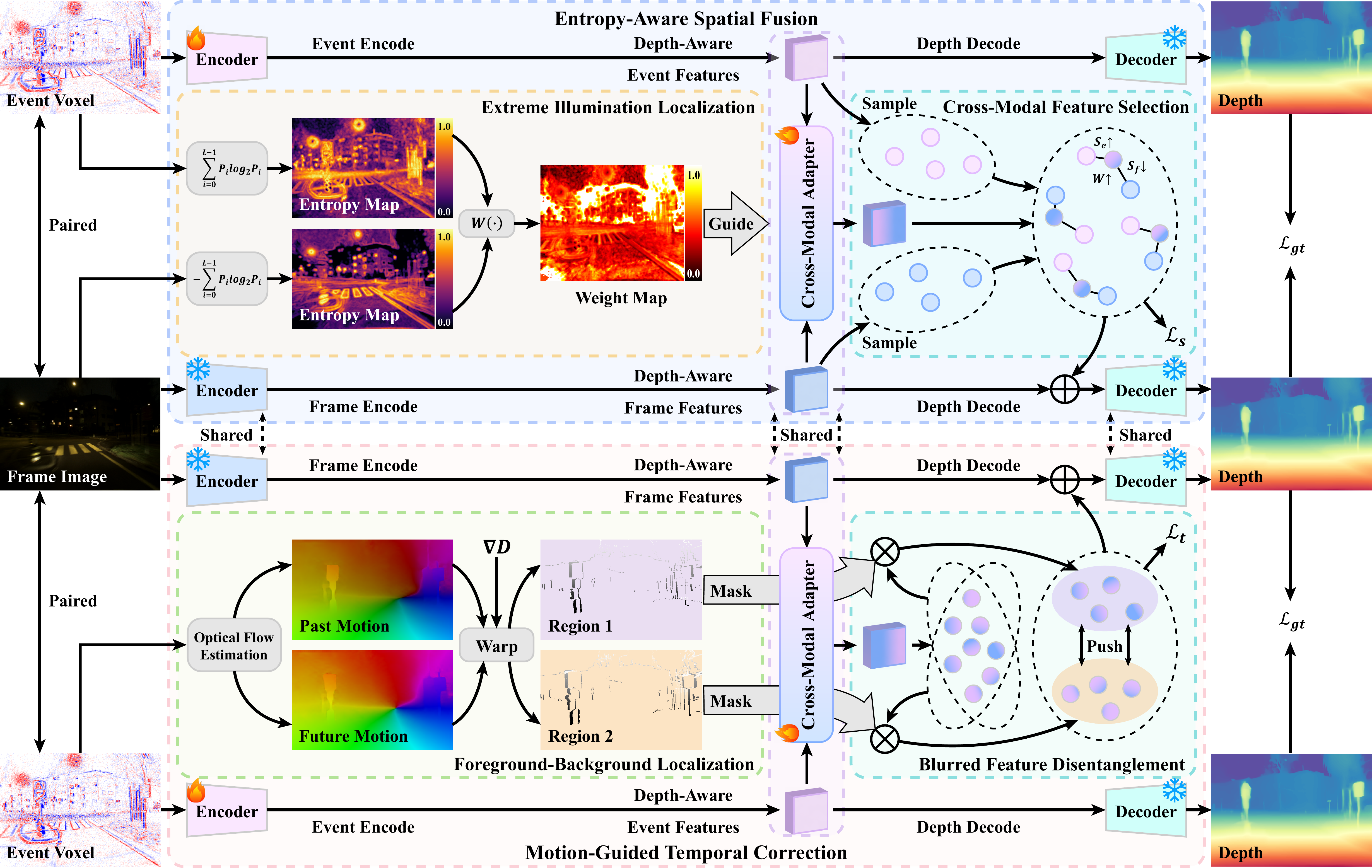}
        \caption
        {   
            Overview of our proposed ADAE framework. ADAE introduces a cross-modal adapter that integrates event features into the frozen Depth Anything model, and further enhances it through Entropy-Aware Spatial Fusion (EASF) and Motion-Guided Temporal Correction (MGTC). The EASF leverages information entropy as a proxy for signal quality to adaptively fuse frame and event features, addressing degradation from extreme illumination. The MGTC utilizes optical flow estimated from events to guide the disentanglement of foreground and background features corrupted by motion blur. 
        }
        \label{Fig:Fusion_Framework}
    \end{figure*}
    
    
    \section{RELATED WORK}
        \subsection{Specialized Models for Depth Estimation}
            Since the first deep learning-based approach \cite{eigen2014depth} for monocular depth estimation, numerous works \cite{patil2022p3depth, fu2018deep, xian2020structure, bhat2021adabins} have introduced various enhancements to improve its performance in normal scenarios. However, these methods exhibit varying degrees of degradation under adverse imaging conditions. Consequently, some approaches \cite{wang2021regularizing, zheng2023steps, vankadari2020unsupervised, liu2021self, saunders2023self, gasperini2023robust, kong2023robodepth} have shifted their focus to depth estimation in extreme environments. These methods often employ GAN-based frameworks \cite{isola2017image, zhu2017unpaired, zheng2020forkgan, pizzati2021comogan} to transfer knowledge learned from clean scenes to degraded scenarios, enabling models to better adapt to various adverse conditions. Nevertheless, these specialized models are typically trained on domain-specific datasets, such as autonomous driving datasets, making it difficult for them to generalize directly to diverse unseen domains effectively.
            
        \subsection{Foundation Models for Depth Estimation}
            To improve generalization across diverse scenes, recent works have developed foundation models for depth estimation \cite{ranftl2020towards, yin2023metric3d, hu2024metric3d, yang2024depth, ke2024repurposing, gui2025depthfm}. Benefiting from training on large-scale hybrid datasets, these models exhibit impressive zero-shot generalization across various scenarios. However, frame-based depth foundation models remain constrained by the imaging limitations of conventional frame cameras, leading to performance degradation under adverse conditions such as extreme illumination and motion blur. Although concurrent work like DA-AC \cite{sun2025depth} attempts to enhance robustness through data augmentation, this approach can only alleviate the problem to a limited extent. It cannot fundamentally overcome the challenge when visual information is irretrievably lost at the sensor level. This highlights that even for the most powerful foundation models, the quality of the input signal remains the ultimate bottleneck.
            
        \subsection{Event-Frame Fusion for Depth Estimation}
            To address the limitations of frame cameras, some methods \cite{gehrig2021combining, shi2023even, xiao2024multimodal, duan2024fusing, devulapally2024multi} introduce event cameras with high dynamic range and high temporal resolution to estimate depth under adverse conditions. These methods leverage the complementary advantages of event and frame to improve depth estimation in extreme environments. However, all these methods require training a complete, specialized network from scratch. This approach not only incurs high computational costs but, more importantly, fails to leverage the vast, generalizable knowledge embedded in modern foundation models like Depth Anything. In contrast, we propose the event-guided spatiotemporal fusion framework ADAE to effectively combine the generalization capabilities of large-scale pre-training with the resilience of event-based sensing.
    
    
    \section{Method}
        \subsection{Framework Overview}
            Our goal is to adapt Depth Anything to adverse imaging conditions, focusing on extreme illumination and motion blur. For frame-based models, extreme illumination leads to spatial information loss, while motion blur blends foreground and background boundaries during exposure. To address these challenges, we inject event data into Depth Anything through a cross-modal adapter with cross-attention mechanisms. As illustrated in Figure \ref{Fig:Fusion_Framework}, the architecture consists of \textit{Entropy-Aware Spatial Fusion} (\textbf{EASF}) and \textit{Motion-Guided Temporal Correction} (\textbf{MGTC}). EASF first employs \textit{Extreme Illumination Localization} to identify degraded regions via information entropy, which then guides \textit{Cross-Modal Feature Selection} to adaptively fuse event and frame features. Similarly, MGTC uses \textit{Foreground-Background Localization} to locate foreground and background regions via dense event-based optical flow, followed by \textit{Blurred Feature Disentanglement} to recalibrate motion-blurred features.

        \subsection{Event Representation}
            To convert the event stream $\{(x_i, y_i, p_i, t_i)\}_{i \in [1, N]}$ into a form suitable for network input, we represent events using voxels. The voxelization process converts events into a tensor $V$ with $B$ bins, as follows \cite{zhu2019unsupervised}: 
            \begin{equation}
                t_i^* = \frac{(B - 1) (t_i - t_1)}{(t_N - t_1)}, 
            \end{equation}
            \begin{equation}
                V(x, y, t) = \sum_i p_i \delta(x,x_i) \delta(y,y_i) max(0, 1 - |t - t_i^*|), 
            \end{equation}
            where $N$ is the number of events, $x_i$ and $y_i$ denote the spatial coordinates of the $i$-th event, while $p_i$ and $t_i$ represent its polarity and timestamp respectively. $\delta(\cdot)$ denotes the Kronecker delta function, which is used to assign events to their corresponding voxel locations.

        \begin{figure*}[t]
            \centering
            \includegraphics[width=1.0\textwidth]{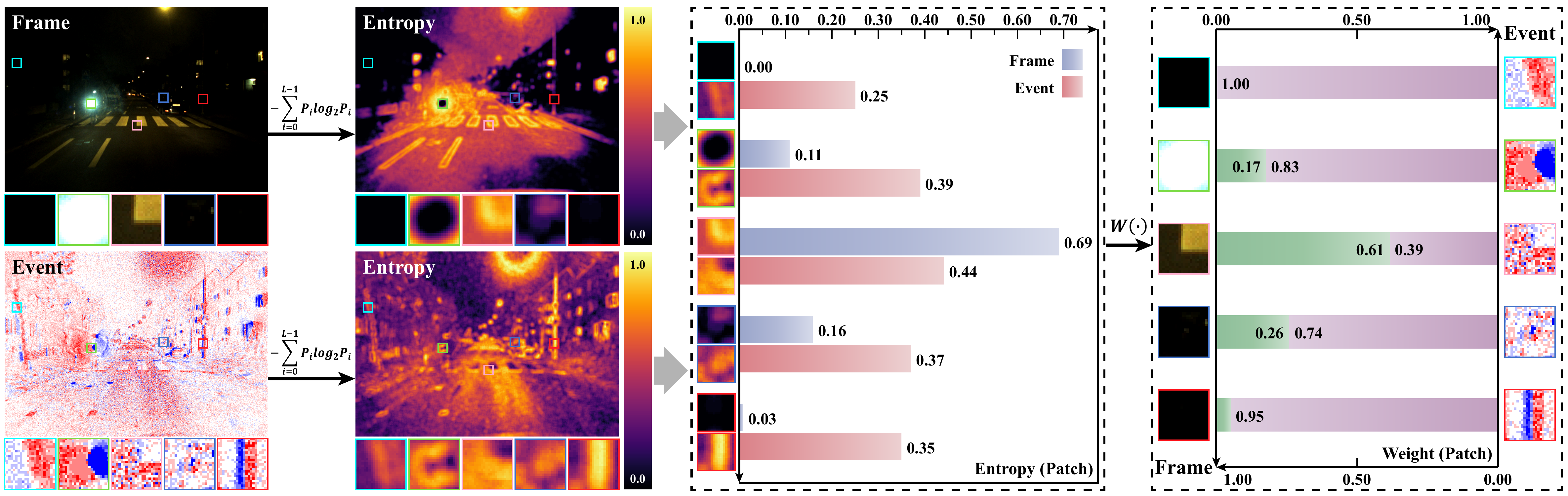}
            \caption
            {
                Motivation of Entropy-Aware Spatial Fusion (EASF). Under extreme illumination, frame suffer from over- or underexposure, while events remain robust but sparse. We observe that information entropy reflects signal reliability in both modalities. This motivates us to adjust fusion weights based on patch-wise entropy comparisons between frame and event modalities, enabling spatially adaptive feature integration under adverse lighting conditions.
            }
            \label{Fig:Spatial_Motivation}
        \end{figure*}

        \subsection{Entropy-Aware Spatial Fusion}
            \subsubsection{Extreme Illumination Localization}
                Frame-based depth estimation suffers from information loss under extreme illumination, while events can provide complementary cues, but their data is inherently sparse. Therefore, naively fusing frame and event features without considering their varying reliability is suboptimal. Figure \ref{Fig:Spatial_Motivation} illustrates that information entropy serves as an effective indicator of signal reliability for both modalities. We computed the information entropy for each local region in both the frame and event voxel, obtaining the corresponding entropy maps $E_f$ and $E_e$. These entropy maps are then used to calculate the weights of each modality according to the following formula:
                \begin{equation}
                    W = \begin{cases}
                            \frac{E_e}{E_e + E_f}, & (E_e + E_f) \geq T \\
                            0.5, & (E_e + E_f) < T
                        \end{cases}, 
                \end{equation}
                where the threshold $T$ assigns a fixed weight of 0.5 to regions where the information entropy of both modalities is relatively low. Note that $E_f$ and $E_e$ are normalized to the range $[0, 1]$, and $W$ and $1-W$ represent the weights of the event and frame modalities, respectively. The weight map subsequently guides the \textit{Cross-Modal Feature Selection}.
                
            \subsubsection{Cross-Modal Feature Selection}
                To enable the cross-modal adapter to adaptively select features from different modalities, we employ the weight map $W$ during training to regulate the distribution similarity between the fused features and the individual modality features. This process can be expressed by the following formulas:
                \begin{equation}
                    S_e = \frac{F_{fused} \cdot F_{e}}{\|F_{fused}\| \cdot \|F_{e}\|}, \text{ } S_f = \frac{F_{fused} \cdot F_{f}}{\|F_{fused}\| \cdot \|F_{f}\|}, 
                \end{equation}
                \begin{equation}
                    \mathcal{L}_{s} = -W\log{\frac{e^{S_e}}{e^{S_e} + e^{S_f}}} - (1 - W)\log{\frac{e^{S_f}}{e^{S_e} + e^{S_f}}}, 
                \end{equation}
                where $F_{fused}$ denotes the fused features produced by the cross-modal adapter, $F_{e}$ and $F_{f}$ represent the features extracted from the event and frozen frame depth encoders, respectively. $S_e$ and $S_f$ denote their cosine similarities, which are constrained by the spatial loss $\mathcal{L}_{s}$. The role of $\mathcal{L}_{s}$ is to reduce the distribution discrepancy between $F_{fused}$ and $F_{e}$ when the weight $W$ is large, and conversely, to reduce the discrepancy between $F_{fused}$ and $F_{f}$ when $W$ is small. This process enables the cross-modal adapter to adaptively select features from different modalities.
        
        \begin{figure*}[t]
            \centering
            \includegraphics[width=1.0\textwidth]{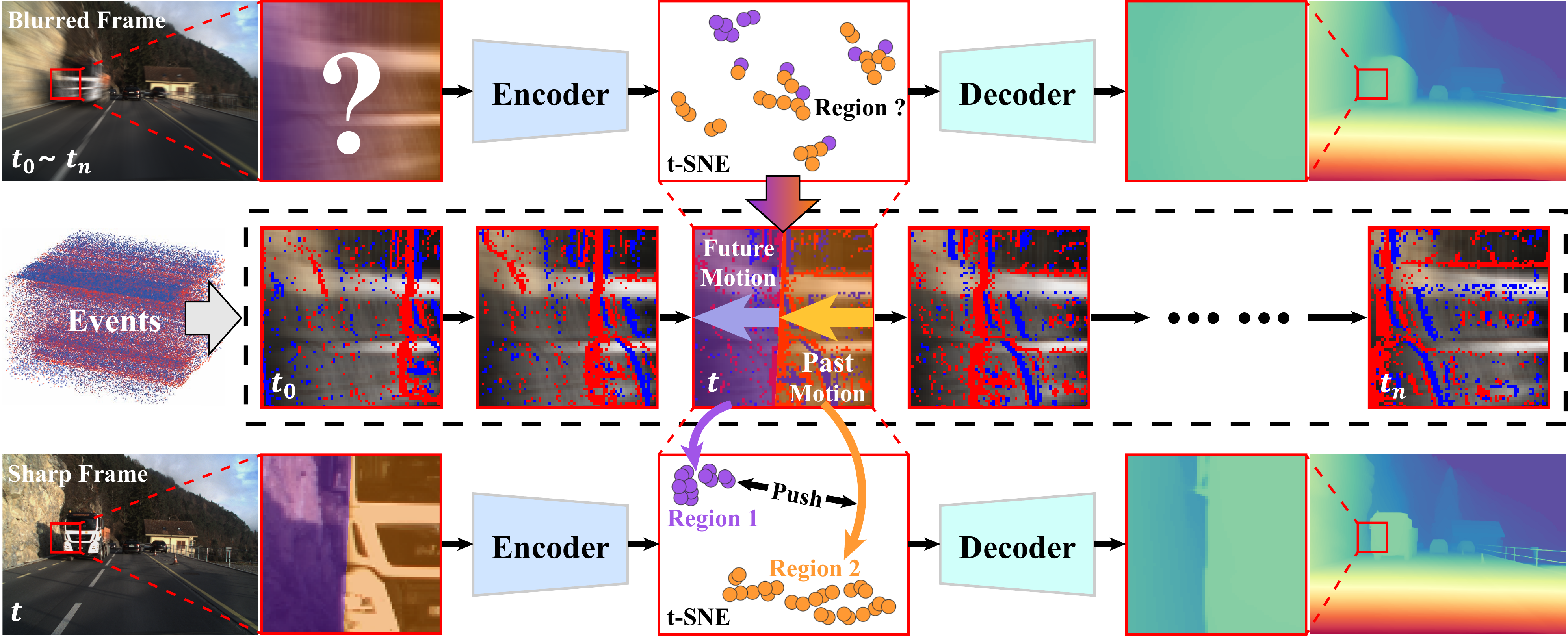}
            \caption
            {
                Motivation of Motion-Guided Temporal Correction (MGTC). We estimate depth on blurred and sharp frames using Depth Anything and visualize their feature distributions via t-SNE. In blurred regions, foreground and background features are entangled, while sharp regions exhibit clear separation. This motivates us to leverage temporally dense event-based optical flow, which captures past and future boundary motions, to localize foreground and background regions within motion-blurred areas. This localization then guides the disentanglement of corrupted features, restoring distinct structural boundaries.
            }
            \label{Fig:Temporal_Motivation}
        \end{figure*}

        \begin{table*}[t]
            \caption{Quantitative results on the DSEC-Degraded dataset. Each image is divided into Normal Illumination Regions and Extreme Illumination Regions to separately evaluate performance under varying illumination conditions, while the Edge Gradient Error (EGE) is computed over the entire image. The best results are highlighted in bold, and the second-best are underlined.}
            \centering
            \setlength{\tabcolsep}{3.9mm}
            \begin{tabular}{l | *{4}{c} | *{4}{c} | c}
                \toprule
                    \multicolumn{1}{c|}{\multirow{2}{*}{\textbf{Methods}}}                                                  & 
                    \multicolumn{4}{c|}{\textbf{Normal Illumination Region}}                                                & 
                    \multicolumn{4}{c|}{\textbf{Extreme Illumination Region}}                                               & 
                    \multicolumn{1}{c}{\multirow{2}{*}{\textbf{EGE} $\boldsymbol{\downarrow}$}}                             \\
                                                                                                                            &
                    \textbf{AbsRel} $\boldsymbol{\downarrow}$                                                               & 
                    $\boldsymbol{\delta_1\uparrow}$ & $\boldsymbol{\delta_2\uparrow}$ & $\boldsymbol{\delta_3\uparrow}$     & 
                    \textbf{AbsRel $\boldsymbol{\downarrow}$}                                                               & 
                    $\boldsymbol{\delta_1\uparrow}$ & $\boldsymbol{\delta_2\uparrow}$ & $\boldsymbol{\delta_3\uparrow}$     \\
                \midrule
                    \textbf{RNW}                                                                                                                                                                       &
                    4.654              & 0.214             & 0.385             & 0.523             & 2.025             & 0.280             & 0.485             & 0.638             & 2.191             \\
                    \textbf{STEPS}                                                                                                                                                                     &
                    4.697              & 0.213             & 0.384             & 0.523             & 2.032             & 0.280             & 0.486             & 0.638             & 2.165             \\
                    \textbf{P3Depth}                                                                                                                                                                   &
                    0.793              & 0.695             & 0.860             & 0.927             & 0.405             & 0.657             & 0.878             & 0.952             & 1.837             \\
                \midrule
                    \textbf{DA-AC}                                                                                                                                                                     &
                    1.163              & 0.809             & 0.915             & 0.951             & 0.459             & 0.808             & 0.940             & 0.973             & 2.530             \\
                    \textbf{DAv2}                                                                                                                                                                      &
                    \textbf{0.402}     & \underline{0.906} & \underline{0.957} & \underline{0.975} & \underline{0.344} & \underline{0.905} & \underline{0.972} & \underline{0.985} & \underline{1.541} \\
                \midrule
                    \textbf{ADAE}                                                                                                                                                                      &
                    \underline{0.409}  & \textbf{0.914}    & \textbf{0.961}    & \textbf{0.978}    & \textbf{0.233}    & \textbf{0.929}    & \textbf{0.981}    & \textbf{0.992}    & \textbf{1.516}    \\
                \bottomrule
            \end{tabular}
            \label{Tab:Relative_Results}
        \end{table*}
        
        \subsection{Motion-Guided Temporal Correction}
            \subsubsection{Foreground-Background Localization}
                As shown in Figure \ref{Fig:Temporal_Motivation}, Depth Anything suffers from deteriorated depth features when estimating depth from motion-blurred frames, leading to corrupted depth structures. This is because the foreground and background boundaries within motion-blurred regions are blended during exposure, weakening the discriminative depth features. To address this issue, we estimate temporally dense event-based optical flow using E-RAFT \cite{gehrig2021raft}, which captures the missing boundary past and future motion within the blurred areas. The estimated flow is then used to warp the depth ground truth gradient $\nabla D$ to different timestamps within the exposure duration, enabling the localization of separated regions corresponding to foreground and background. These regions are represented as masks, which guide the following \textit{Blurred Feature Disentanglement}. Note that $\nabla D$ is employed to suppress interference from texture-induced edges.
            
            \subsubsection{Blurred Feature Disentanglement}
                After obtaining the foreground-background masks, we adopt a supervised contrastive loss \cite{khosla2020supervised} to encourage intra-class feature compactness and inter-class feature separation:
                \begin{equation}
                    \mathcal{L}_{t} = - \frac{1}{N} \sum_{i=1}^{N} \frac{1}{|P(i)|} \sum_{j \in P(i)} \log \frac{\exp \left( \frac{f_i \cdot f_j}{\tau} \right)}{\sum_{\substack{k=1 \\ k \neq i}}^{N} \exp \left( \frac{f_i \cdot f_k}{\tau} \right)}, 
                \end{equation}
                where $f_i$ denotes the normalized feature vector at pixel $i$, $P(i)$ is the set of positive samples that share the same foreground or background label with $i$, $N$ is the total number of valid features, and $\tau$ is the temperature scaling factor.

        \subsection{Training Details}
            \subsubsection{Optimization} 
                The overall loss of the proposed framework is formulated as follows: 
                \begin{equation}
                    \mathcal{L}_{ADAE} = \lambda_1 \mathcal{L}_{gt} + \lambda_2 \mathcal{L}_{s} + \lambda_3 \mathcal{L}_{t}, 
                \end{equation}
                where $\lambda_1$, $\lambda_2$ and $\lambda_3$ denote the weights assigned to balance the influence of each loss component. $\mathcal{L}_{gt}$ is the scale-invariant loss \cite{eigen2014depth}: 
                \begin{equation}
                    d_i = \log{D_i} - \log{D_i^*}, 
                \end{equation}
                \begin{equation}
                    \mathcal{L}_{gt} = \frac{1}{N} \sum_{i=1}^N{d_i^2} - \frac{1}{N^2} (\sum_{i=1}^N{d_i})^2, 
                \end{equation}
                where $D_i$ denotes the network output, and $D_i^*$ represents the depth ground truth.
            
            \subsubsection{Implementation} 
                We train the model in two steps. In the first step, the event encoder is pre-trained by optimizing the following loss function:
                \begin{equation}
                    \mathcal{L}_{pretrain} = \frac{1}{N} \left\| F_{f} - F_{e} \right\|_{1}, 
                \end{equation}
                where $\mathcal{L}_{pretrain}$ facilitates the event encoder in acquiring basic knowledge from the frozen frame encoder. In the second step, we trained the model for 100 epochs using the AdamW optimizer with an initial learning rate of 0.0001. The weights $\lambda_1$, $\lambda_2$ and $\lambda_3$ were set to 1.0, 0.2, and 0.1, respectively. All training was conducted using PyTorch on 6 NVIDIA RTX 5090 GPUs. Note that during testing, the final model only consists of the Depth Anything model, the cross-modal adapter, and the event encoder.
    
    
    \section{Experiments}
        \subsection{Experimental Settings}
            \subsubsection{Datasets}
                We conduct experiments on DSEC \cite{gehrig2021dsec}, EVRB \cite{kim2024cmta}, NCER \cite{cho2023non}, RELED \cite{kim2024towards}, and MVSEC \cite{zhu2018multivehicle} datasets, all of which provide event modalities. The DSEC dataset is used to synthesize DSEC-Degraded, which features extreme illumination and motion blur. Specifically, we first apply the event-based video frame interpolation method \cite{ding2024video} to upsample the DSEC images for motion blur synthesis. Subsequently, we simulate extreme illumination conditions using the following equation:
                \begin{equation}
                    I_{out} = 0.5 + \alpha (I_{in} - 0.5), 
                \end{equation}
                where $I_{in}$ and $I_{out}$ denote the input and output pixel intensities normalized to $[0, 1]$, and $\alpha$ is a stretching factor that amplifies contrast towards over- or underexposure. The resulting values are clipped to $[0, 1]$. The depth annotations of DSEC-Degraded are generated by Depth Anything inference on the original DSEC images. Furthermore, we manually select samples with real motion blur from the event-frame deblurring datasets EVRB, NCER, and RELED to assess the zero-shot generalization capability of our method. Their depth ground truth is obtained via Depth Anything on the corresponding clean images. Additionally, we compare our method with existing event-frame fusion approaches on MVSEC, which contains real-world extreme illumination scenarios, for metric depth estimation.

    \begin{figure*}[t]
        \centering
        \includegraphics[width=1.0\textwidth]{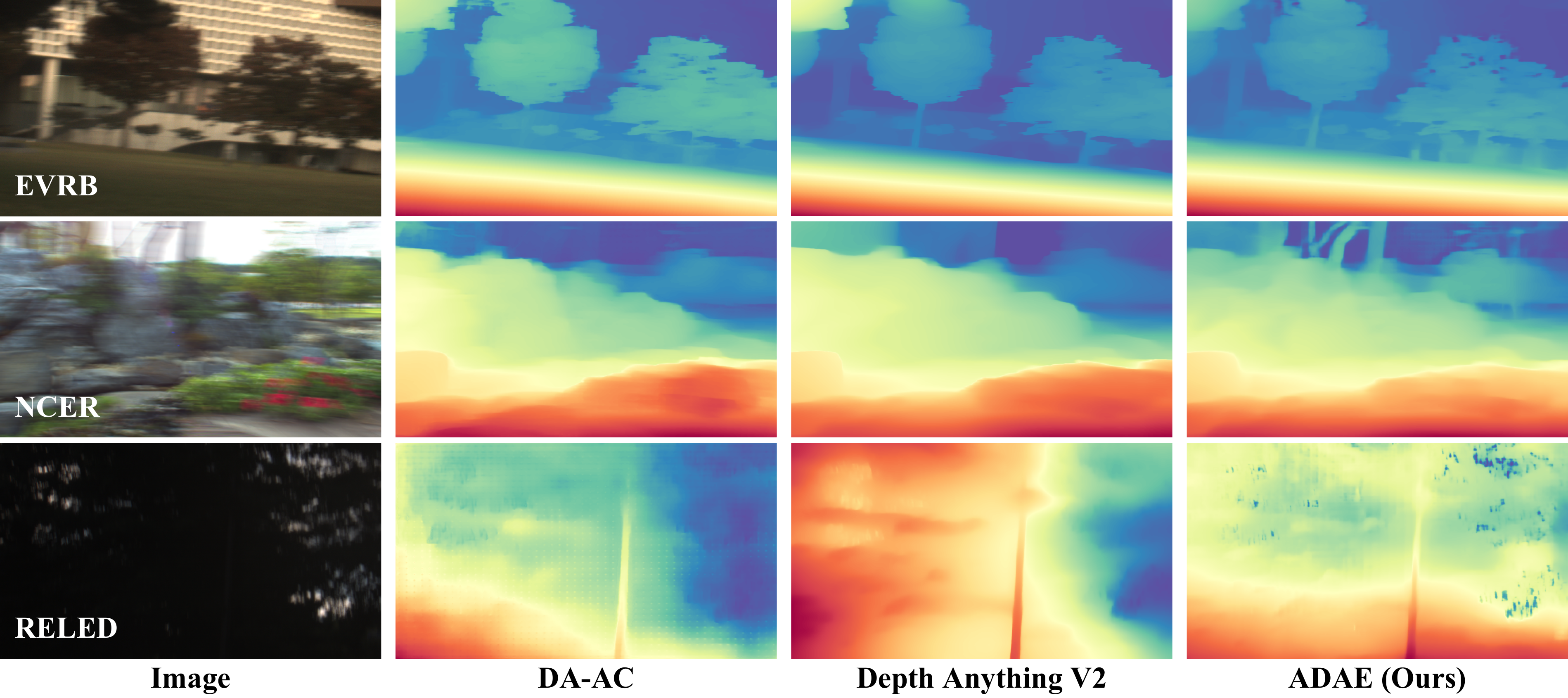}
        \caption
        {
            Zero-shot qualitative comparison in real-world adverse imaging conditions.
        }
        \label{Fig:Qualitative_Comparison}
    \end{figure*}

    \begin{table*}[t]
        \caption{Zero-shot quantitative comparison in real-world adverse imaging conditions.}
        \centering
        \setlength{\tabcolsep}{1.70mm}
        \begin{tabular}{l | *{4}{c} | *{4}{c} | *{4}{c}}
            \toprule
                \multicolumn{1}{c|}{\multirow{2}{*}{\textbf{Methods}}}                                                               & 
                \multicolumn{4}{c|}{\textbf{EVRB}}                                                                                   & 
                \multicolumn{4}{c|}{\textbf{NCER}}                                                                                   &
                \multicolumn{4}{c}{\textbf{RELED}}                                                                                   \\
                                                                                                                                     & 
                \textbf{AbsRel} $\boldsymbol{\downarrow}$ & $\boldsymbol{\delta_1\uparrow}$ & $\boldsymbol{\delta_2\uparrow}$ & \textbf{EGE} $\boldsymbol{\downarrow}$ &
                \textbf{AbsRel} $\boldsymbol{\downarrow}$ & $\boldsymbol{\delta_1\uparrow}$ & $\boldsymbol{\delta_2\uparrow}$ & \textbf{EGE} $\boldsymbol{\downarrow}$ &
                \textbf{AbsRel} $\boldsymbol{\downarrow}$ & $\boldsymbol{\delta_1\uparrow}$ & $\boldsymbol{\delta_2\uparrow}$ & \textbf{EGE} $\boldsymbol{\downarrow}$ \\
            \midrule
                \textbf{DA-AC}                                                                                                                                          & 
                1.580             & 0.214             & 0.402             & 1.860             &
                0.856             & 0.485             & 0.706             & 1.880             &
                1.543             & 0.333             & 0.541             & 1.899             \\
                \textbf{DAv2}                                                                                                                                           & 
                \underline{0.969} & \textbf{0.364}    & \underline{0.611} & \textbf{1.705}    &
                \textbf{0.360}    & \underline{0.634} & \underline{0.832} & \underline{1.493} &
                \underline{0.815} & \underline{0.421} & \underline{0.679} & \textbf{1.737}    \\
                \textbf{ADAE}                                                                                                                                           & 
                \textbf{0.792}    & \underline{0.350} & \textbf{0.623}    & \underline{1.750} &
                \underline{0.395} & \textbf{0.700}    & \textbf{0.837}    & \textbf{1.489}    & 
                \textbf{0.676}    & \textbf{0.486}    & \textbf{0.749}    & \underline{1.775} \\
            \bottomrule
        \end{tabular}
        \label{Tab:Zeroshot_Datasets}
    \end{table*}

    \begin{table}[t]
        \caption{Quantitative results on the MVSEC dataset.}
        \centering
        \setlength{\tabcolsep}{3.4mm}
        \begin{tabular}{l | *{4}{c}}
            \toprule
                \multicolumn{1}{c|}{\textbf{Methods}} & \textbf{day1}    & \textbf{night1}  & \textbf{night2}  & \textbf{night3}  \\
            \midrule
                \textbf{RAMNet}             & 2.76             & 3.82             & 3.28             & 3.43             \\
                \textbf{ER-F2D}             & 2.62             & \textbf{2.78}    & \underline{2.95} & \underline{2.81} \\
                \textbf{SRFNet}             & \underline{2.37} & 3.01             & 3.22             & 3.52             \\
                \textbf{ADAE}               & \textbf{2.26}    & \underline{2.89} & \textbf{2.77}    & \textbf{2.75}    \\
            \bottomrule
        \end{tabular}
        \label{Tab:Absolute_Results}
    \end{table}

            \subsubsection{Comparison Methods and Metrics} 
                We compare our model with frame-based specialized models: RNW \cite{wang2021regularizing}, STEPS \cite{zheng2023steps}, and P3Depth \cite{patil2022p3depth}, frame-based foundation models: DA-AC \cite{sun2025depth} and Depth Anything V2 (DAv2) \cite{yang2024depth}, and event-frame fusion specialized models: RAMNet \cite{gehrig2021combining}, ER-F2D \cite{devulapally2024multi}, and SRFNet \cite{pan2024srfnet}. We use Absolute Relative Error (AbsRel) and accuracy under threshold ($\delta_i$) as evaluation metrics, and further employ an Edge Gradient Error (EGE) to evaluate the model's performance at depth boundaries: 
                \begin{equation}
                    \text{EGE} = \frac{1}{N_G} \sum_{i=1}^{N} \mathbb{I} \left( |\nabla D_i^*| > G \right) \cdot \frac{|\nabla D_i - \nabla D_i^*|}{|\nabla D_i^*|}, 
                \end{equation}
                where $\nabla D_i$ and $\nabla D_i^*$ denote the gradients of the predicted and ground truth depth at pixel $i$, respectively, and $\mathbb{I}(\cdot)$ denotes the indicator function that equals $1$ if the condition is satisfied and $0$ otherwise. $G$ is a threshold used to select significant depth edges. In addition, following \cite{gehrig2021combining}, we use the Mean Absolute Error (MAE) for metric depth evaluation.

    \begin{table}[t]
        \caption{Ablation study on key components of ADAE.}
        \centering
        \setlength{\tabcolsep}{1.5mm}
        \begin{tabular}{cc|ccccc}
            \toprule
                $\textbf{EASF}$ & $\textbf{MGTC}$                                                                                                   & 
                \textbf{AbsRel $\boldsymbol{\downarrow}$}                                                                                           & 
                $\boldsymbol{\delta_1\uparrow}$ & $\boldsymbol{\delta_2\uparrow}$ & $\boldsymbol{\delta_3\uparrow}$                                 &
                \textbf{EGE} $\boldsymbol{\downarrow}$                                                                                              \\
            \midrule
                \usym{2613} & \usym{2613} & 0.1466             & 0.9119             & 0.9632             & \underline{0.9805} & 1.5487              \\
                \faCheck    & \usym{2613} & \underline{0.1454} & \underline{0.9143} & \underline{0.9659} & \textbf{0.9824}    & 1.5580              \\
                \usym{2613} & \faCheck    & 0.1459             & 0.9135             & 0.9628             & 0.9804             & \underline{1.5214}  \\
                \faCheck    & \faCheck    & \textbf{0.1444}    & \textbf{0.9164}    & \textbf{0.9667}    & \textbf{0.9824}    & \textbf{1.5155}     \\
            \bottomrule
        \end{tabular}
        \label{Tab:Ablation_Modules}
    \end{table}

    \begin{table}[t]
        \caption{Ablation study on different adapter capacities.}
        \centering
        \setlength{\tabcolsep}{1.4mm}
        \begin{tabular}{l | *{2}{c} | *{4}{c}}
            \toprule
                \textbf{Adapter}                                                                                                        & 
                \textbf{Params} & \textbf{Runtime}                                                                                      &
                \textbf{AbsRel $\boldsymbol{\downarrow}$} & $\boldsymbol{\delta_1\uparrow}$ & \textbf{EGE} $\boldsymbol{\downarrow}$    \\
            \midrule
                \textbf{Small}  & 25.215 M & 101.89 ms & 0.1697             & 0.9084                & 1.5448                \\
                \textbf{Medium} & 33.608 M & 102.34 ms & 0.1479             & 0.9133                & \underline{1.5400}    \\
                \textbf{Large}  & 50.393 M & 103.73 ms & \underline{0.1453} & \underline{0.9161}    & 1.5445                \\
            \midrule
                \textbf{ADAE}   & 42.000 M & 103.29 ms & \textbf{0.1444}    & \textbf{0.9164}       & \textbf{1.5155}       \\
            \bottomrule
        \end{tabular}
        \label{Tab:Ablation_Adapters}
    \end{table}
        
        \subsection{Comparison Experiments}
            \subsubsection{Comparison under Synthetic Adverse Conditions} 
                In Table \ref{Tab:Relative_Results}, we compare various methods on the DSEC-Degraded dataset with synthetic adverse imaging conditions. Frame-based specialized models are limited by their inherent capacity and thus perform worse than frame-based foundation models. Nevertheless, even foundation models suffer from information loss and depth structural distortions under extreme illumination and motion blur. In contrast, our method demonstrates robust performance in both normal and extreme illumination regions, effectively leveraging event modalities to enhance degraded areas while maintaining accuracy in well-exposed regions. Our method also achieves the lowest EGE, validating its effectiveness in addressing depth distortions caused by motion blur.

            \subsubsection{Comparison under Real-World Adverse Conditions} 
                The quantitative and qualitative comparisons on unseen real-world challenging scenarios are presented in Table \ref{Tab:Zeroshot_Datasets} and Figure \ref{Fig:Qualitative_Comparison}. The proposed method outperforms other approaches on most metrics, demonstrating its zero-shot generalization capability. Furthermore, we evaluate the performance of our method on the event-frame fusion metric depth estimation using the MVSEC dataset. Following previous works \cite{gehrig2021combining, devulapally2024multi, pan2024srfnet}, we fine-tune only on the \textit{day2} sequence. Table \ref{Tab:Absolute_Results} presents the MAE evaluated within 30 meters, demonstrating the competitive performance of our method under real-world challenging imaging conditions.
         
        \subsection{Ablation Study}
            \subsubsection{Effectiveness of Key Components in ADAE}
                Table \ref{Tab:Ablation_Modules} presents the ablation study on the \textit{Entropy-Aware Spatial Fusion} (\textbf{EASF}) and \textit{Motion-Guided Temporal Correction} (\textbf{MGTC}) on the DSEC-Degraded dataset. EASF improves overall depth estimation performance, evidenced by reductions in AbsRel and improvements in $\delta_i$ metrics. On the other hand, MGTC is particularly effective in enhancing the model's capability to recover structural details in motion-blurred regions, as indicated by the decrease in the EGE. The model achieves the best results when both modules are combined, demonstrating their complementary contributions.
            
            \subsubsection{Influence of Adapter Capacity}
                Table \ref{Tab:Ablation_Adapters} investigates the relationship between adapter capacity and model performance. All runtime measurements were conducted on a single NVIDIA RTX 5090 GPU. As observed, increasing the adapter capacity initially yields performance gains. However, as the capacity scales up, the performance improvement on the validation set becomes marginal or even suffers from potential overfitting. Therefore, instead of simply increasing the adapter capacity, we selected a moderate capacity that strikes a balance between performance and efficiency.
    
    
    \section{Conclusion}
        In this work, we presented ADAE, an event-guided spatiotemporal fusion framework that enhances the robustness of Depth Anything under extreme illumination and motion blur. Our approach leverages the complementary properties of events and frames by introducing a cross-modal adapter that integrates event signals into the frozen depth foundation model. To address illumination degradation, we proposed \textit{Entropy-Aware Spatial Fusion}, which adaptively adjusts fusion weights based on patch-wise entropy. To correct motion-induced feature ambiguity, we introduced \textit{Motion-Guided Temporal Correction}, which leverages temporally dense event-based optical flow to restore foreground-background boundaries. Extensive experiments across multiple datasets demonstrate that ADAE improves depth estimation in adverse imaging conditions while preserving the generalization capability of the depth foundation model. In the future, we plan to explore more efficient event representations and extend our event-enhanced fusion framework to broader pixel-level perception tasks in challenging environments.

    
    \addtolength{\textheight}{-6cm}
    \bibliographystyle{IEEEtranBST/IEEEtran}
    \bibliography{references}
\end{document}